\documentclass[runningheads]{llncs}
\usepackage[T1]{fontenc}
\usepackage{graphicx}
\usepackage{booktabs}
\usepackage[misc]{ifsym}

\usepackage{mwe}
\graphicspath{{figures/}}
\usepackage[ruled,vlined]{algorithm2e}
\usepackage{paralist}
\usepackage{color}
\usepackage{amsmath}
\usepackage{amsfonts}
\usepackage[colorlinks=true, allcolors=blue]{hyperref}
\usepackage{bm}
\usepackage{comment}
\usepackage{multicol}
\usepackage{subcaption}
\usepackage{fontawesome5}
\usepackage[utf8]{inputenc}
\usepackage{tabularx}
\usepackage{float}
\usepackage{algpseudocode}
\usepackage{threeparttable} 
\usepackage{soul}
\usepackage{booktabs} % Add booktabs

\usepackage{booktabs}
\usepackage{multirow}
\usepackage[table]{xcolor}
\usepackage{xcolor}

\usepackage{multirow}
\usepackage{booktabs}
\usepackage{array}

\newcommand{\remove}[1]{}   % camera ready

\begin{document}

% \title{Stochastic real-time planning with Monte Carlo rollouts to boost RL performance for a real world optimization problem}

% \title{Curriculum RL Meets Monte Carlo Planning: Tackling Collisions in a Real World Container Management Problem}

\title{Curriculum RL meets Monte Carlo Planning: Optimization of a Real World Container Management Problem}

\titlerunning{Curriculum RL meets Monte Carlo Planning}
% If the full title of your paper is short enough to also fit in the running head, you can omit the abbreviated paper title here. You can check as follows: if you comment out the \titlerunning line, something will appear in the header of all odd-numbered pages of your PDF from page 3 onward. This is something is either the full title (in which case all is well), or the error message "Title Suppressed Due to Excessive Length". If this error message appears, you're going to want to provide an abbreviated title within the \titlerunning command, because if you don't do it, Springer will do it for you.

%N.B.: Author information (both in the \author{} and \authorrunning{} command) should only be present in the Camera-Ready Version of your paper. The version that you initially submit for review, ought to be double-blind. So, when initially submitting your paper, use:
%\author{Author information scrubbed for double-blind reviewing}

\author{Author\inst{1}\faIcon{envelope} \and Author\inst{1}}

%\author{Abhijeet Pendyala \corr \and Asma Atamna \and Tobias Glasmachers}
% You may leave out the orcidID information, if you want to.
% Use \corr to indicate the corresponding author. Note the spacing around the \corr command. Only one author can be the corresponding author.

%N.B.: comment out the \authorrunning{} command for the double-blind version of your paper submitted for review. Later, if your paper is accepted, use the command for the Camera-Ready Version.

%\authorrunning{anonymous author(s)}

% If there are two authors, write 'A.L. Benjamin and C.C. Broadus Jr.'
% If there are more than two authors, '[...] et al.' is used.

\author{Abhijeet Pendyala \faIcon{envelope} \and
Tobias Glasmachers
}

\authorrunning{A. Pendyala et al.}

\institute{Ruhr-University Bochum, Bochum, Germany\\
\email{firstname.lastname@ini.rub.de}
}

\maketitle

\begin{abstract}

In this work, we augment reinforcement learning with an inference-time collision model to ensure safe and efficient container management in a waste-sorting facility with limited processing capacity. Each container has two optimal emptying volumes that trade off higher throughput against overflow risk. Conventional reinforcement learning (RL) approaches struggle under delayed rewards, sparse critical events, and high-dimensional uncertainty—failing to consistently balance higher-volume empties with the risk of safety-limit violations. To address these challenges, we propose a hybrid method comprising: (1) a \emph{curriculum-learning} pipeline that incrementally trains a PPO agent to handle delayed rewards and class imbalance, and (2) an \emph{offline pairwise collision model} used at inference time to proactively avert collisions with minimal online cost. Experimental results show that our \emph{targeted inference-time collision checks} significantly improve collision avoidance, reduce safety-limit violations, maintain high throughput, and scale effectively across varying container-to-PU ratios. These findings offer actionable guidelines for designing safe and efficient container-management systems in real-world facilities.

\end{abstract}

\section{Introduction}\label{sec:intro}

Waste-sorting facilities increasingly rely on data-driven methods to meet strict energy efficiency and sustainability goals, due in part to regulatory directives for responsible recycling of packaging waste. Modern plants must deal with fluctuating material types, unpredictable daily volumes, and stringent safety constraints. This work is inspired by the final stage of a waste-sorting facility where \emph{n} containers (\emph{bunkers}) accumulate various types of material at unique stochastic rates. These materials are subsequently transported to a processing unit (PU) for compaction into bales or products \footnote{In this study the terms bunker/container and processing unit (PU)/press are used interchangeably.} during which PU is unavailable for further processing. In addition, each container has a strict maximum capacity and overflowing beyond a threshold requires halting the facility for corrective measures, incurring steep penalties. In this context, \emph{container management} emerges as a critical bottleneck: Emptying these containers too late risks overflow; emptying them too early or too frequently undermines throughput and raises energy costs. 

Recent work has modeled container management as a reinforcement learning (RL) problem, notably through \emph{ContainerGym}~\cite{Pendyala2024-ContainerGym}, providing a real-world benchmark where an RL agent decides \emph{when} to empty each container. However, naive Proximal Policy Optimization (PPO) agents frequently fail in this domain, as \emph{delayed rewards}, \emph{sparse critical events}, and a \emph{single shared PU} create complex scheduling dynamics. For instance, some containers have a ``higher peak'' volume (around 60–75\% capacity) for optimal throughput and a ``lower peak'' (around 30–40\%) as a fallback. Emptying containers at the peak volumes yields optimal products, hence emptying at other volumes is undesirable. The prime reason for missing a peak volume is unavailability of the PU caused by a \emph{collision} of two or more containers reaching peak volume around the same time. Prior work showed that \emph{curriculum learning} helps mitigate early overflows and improves PPO’s learning curve~\cite{Pendyala2024-Solving}. Still, the problem of handling collisions remains unsolved, especially at higher container-to-PU ratios.

On the other hand, despite the installation of sophisticated sensors in these facilities and real-time monitoring, most waste sorting design layouts remain fixed once built, even if data points to major bottlenecks. This reveals a broader gap: leveraging RL not only to optimize day-to-day operations but also to guide \emph{facility design decisions} such as how many containers can safely share a PU before collisions dominate. In other industries—like robotics or warehouse logistics—RL insights have influenced layout reconfiguration, helping systems adapt hardware choices to software-derived constraints. Yet in waste-sorting, this feedback loop remains largely unexplored.

To address these gaps, we present a \emph{hybrid RL} method that integrates curriculum learning with a domain-specific \emph{collision model} at inference time. Our approach (1) reduces collision-induced overflows and throughput losses, and (2) closes the loop between software-driven scheduling and hardware design insights. Specifically, we:
\begin{itemize}
    \item Propose a three-phase \emph{curriculum learning} strategy to tackle delayed rewards and dual-volume targets (higher vs.\ lower peak).
    \item Introduce an \emph{offline-trained collision model} for on-the-fly inference checks, overriding risky “no-op” actions when multiple containers approach critical volumes.
    \item Systematically evaluate varying \emph{container-to-PU ratios} (7:1 to 12:1), providing actionable guidelines on how many containers a single PU can realistically handle without causing excessive collisions.
\end{itemize}

Empirical results demonstrate that the proposed hybrid RL framework significantly cuts collisions, reduces safety limit violations, maintains higher throughput, and yields design insights for scaling container management. 

% The remainder of this paper is organized as follows. Section~\ref{sec:related_work} situates our approach within the broader literature on industrial RL, curriculum learning, and inference-time planning. Section~\ref{sec:real_env_description} formalizes the container-management environment, while Section~\ref{sec:methodology} details our training pipeline. Section~\ref{sec:Collision_Model_and_Inference_Pipeline} outlines the collision model integration, and Section~\ref{sec:results} presents empirical evaluations. Finally, Section~\ref{sec:conclusion} offers conclusions and avenues for future research.

\begin{figure}
    \begin{center}
        \includegraphics[clip, trim=0.1cm 0.3cm 0.1cm 0.1cm,width=0.95\textwidth]{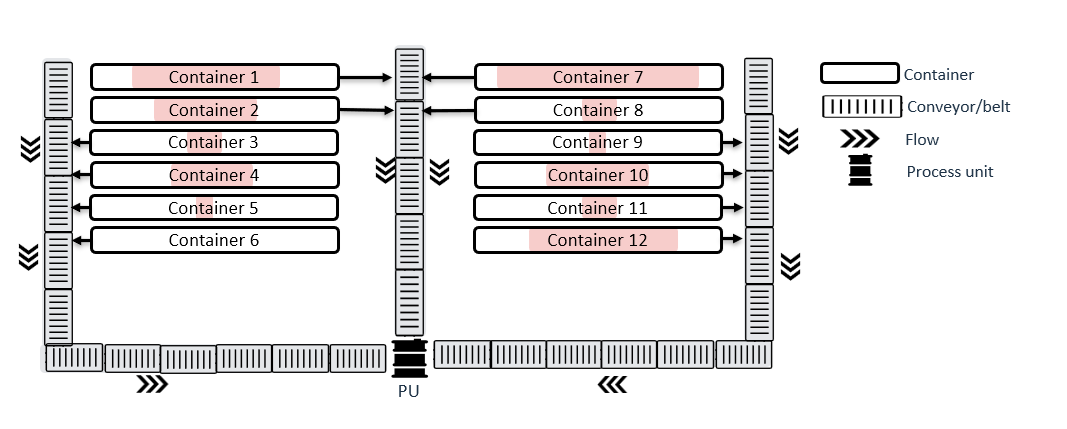}
    \end{center}
    \caption{\label{figure:schematic}
        Layout sketch of a facility with 12 containers and a PU, connected with conveyor belts. The containers are filled from above, with their current fill states indicated by the shaded areas.
    }
\end{figure}

\section{Related Work}\label{sec:related_work}

Reinforcement learning has proliferated in industrial and logistics applications, moving beyond canonical benchmarks to real-world applications. We categorize related literature under the following three key themes: 

\paragraph{\textbf{Curriculum Learning in Industrial RL}}
{Curriculum learning (CL)} systematically organizes the training tasks, typically starting with simpler sub-problems before moving to more challenging ones~\cite{Narvekar2020-fn,Wang2022-dc}. In industrial domains, where critical actions are rare and reward signals can be delayed or sparse, CL helps agents gather meaningful experience without being overwhelmed by complex dynamics from the outset. Prior work in container management~\cite{Pendyala2024-Solving} showed that a curriculum-based PPO significantly reduces early overflows compared to naive baselines. We build on this by designing a multi-phase reward curriculum—enabling the agent to master \emph{dual-peak} emptying targets and maintain stable performance under varying inflows.

\paragraph{\textbf{Inference-Time Planning and Collision Avoidance}}
While a trained RL policy provides baseline decisions at deployment, \emph{inference-time planning} augments these decisions with lookahead logic or heuristic checks, often through Monte Carlo methods. Techniques vary in whether they update the agent’s parameters or remain “stateless.” Prominent examples include Monte Carlo simulations in game-playing AI~\cite{Silver2016,Browne2012}, but similar ideas have surfaced in robotics~\cite{Romero2023-MPC,Wang2022}, energy systems control~\cite{Eseye2022} and autonomous driving~\cite{Hoel2020}. In our context, we adopt a \emph{collision model} that is trained offline to predict when multiple containers are poised to exceed safe volumes simultaneously. This model supplements a curriculum-trained policy by overriding risky no-operation actions—preventing multi-container collisions. By decoupling real-time safety checks from the offline-trained policy, our method maintains policy stability and adds minimal inference overhead, which is essential for deployment in high-throughput industrial environments.

\paragraph{\textbf{RL-Driven Design insights}}
An emerging trend in industrial settings is the use of RL not merely to control a dynamic process but also to inform insights into \emph{design} decisions. In large warehouse environments, multi-agent RL has optimized the assignment of ``chutes'' to destinations~\cite{Shen2023}, reducing robot congestion and improving throughput. Similar approaches in factory automation focus on workstation placement~\cite{Ikeda2022}, where RL-driven layout designs outperform handcrafted alternatives. Meanwhile, open-source platforms like ``Storehouse'' show that RL can outperform heuristic policies in dynamic warehouse slotting~\cite{Storehouse2022}. These successes underscore how AI insights can \emph{redesign} processes for higher efficiency, paralleling our goal of using RL not just to operate container management but to shape decisions on how many containers a PU can handle effectively.

\paragraph{\textbf{Context of our Contribution}}
By integrating a \emph{predefined curriculum} for delayed rewards with an \emph{offline collision predictor}, our hybrid \textbf{PPO-CL-CM} approach targets both throughput optimization and collision avoidance in container management. Offline pairwise simulations yield a lightweight collision classifier, which can be queried at each time step to override the policy when collision risk is high. The result is a system that meets key challenges—dual-peak scheduling, sparse rewards, and collision hazards—while also generating real-world design insights. In the following sections, we present the formal environment setup, detail our curriculum learning phases, and then describe how we integrate collision checks at inference.

% In the following sections, we present the formal environment setup, detail our curriculum learning phases, and then describe how we integrate collision checks at inference. We also provide extensive evaluations showing that this framework outperforms naive PPO and plain curriculum RL in both safety metrics and long-term throughput, especially as the number of containers grows.

% \begin{figure}[h]
%     \centering
%     \begin{minipage}{0.6\textwidth}
%         \centering
%         \includegraphics[clip, trim=0.1cm 0.3cm 0.1cm 0.1cm, width=\textwidth]{schematic.png}
%         \caption{Layout sketch of a facility with 12 containers and a PUs, connected with conveyor belts.}
%         \label{figure:schematic}
%     \end{minipage}
% \end{figure}

\section{Environment and RL formulation}\label{sec:real_env_description}

% This work is inspired by the final stage of a waste-sorting facility where \emph{n} containers (\emph{bunkers}) accumulate various types of material at unique stochastic rates. These materials are subsequently transported to a \emph{single} processing unit (PU) for compaction into bales or products \footnote{In this study the terms bunker/container and processing unit (PU)/press are used interchangeably.}. Containers must be emptied completely, and the PU is unavailable during processing. In addition, each container has a strict maximum capacity of 40 units and overflowing beyond this threshold requires halting the facility for corrective measures, incurring steep penalties. 

In this section, we provide details of the considered container management environment.
Building on the scenario described in Section~\ref{sec:intro}, we reiterate the central design optimization criterion; each container has two preferred or ``ideal'' volumes at which emptying yields the highest-quality output. These correspond to a \emph{Higher peak} offering better overall throughput if the container can safely reach this point and a \emph{Lower peak}, providing a reasonable fallback when waiting longer could risk overflow or collide with the PU's availability window. In practice, larger volumes generally improve efficiency, as the PU operates more effectively when processing bigger batches at once. However, strictly aiming for the higher peak can provoke collisions or safety limit breaches if the single PU is not available in time.

\paragraph{Markov Decision Process (MDP) Setup.}
We model the container-management scenario as an MDP $(\mathcal{S}, \mathcal{A}, p, r, \gamma)$:
\begin{compactitem}
    \item \textbf{State $s_t$:} Comprises volumes $\{v_{i,t}\}_{i=1}^n$ for each container, a PU-availability counter $p_t$ (time until the processing unit becomes available), and auxiliary signals such as ideal volumes. This provides the agent with both physical constraints (capacity, current usage) and strategic cues (optimal targets).
    \item \textbf{Action $a_t \in \{0,\dots,n\}$:} Either do nothing ($a_t=0$) or attempt to empty container $i$. If $p_t>0$ (the PU is busy), an emptying request fails, and container $i$ continues to fill. This design highlights collision risks when multiple containers approach peak volumes simultaneously.
    \item \textbf{Volume Dynamics:} Each container $i$ grows according to a \emph{random walk with drift}:
    \begin{equation}\label{eq:random_walk}
    v_{i,t+1} \;=\; \max\!\bigl(0,\;v_{i,t} + \alpha_i + \epsilon_{i,t}\bigr),
    \end{equation}
    where $\alpha_i$ is the average fill rate, and $\epsilon_{i,t}\!\sim\!\mathcal{N}(0,\sigma_i^2)$ captures stochastic fluctuations. Overflow occurs if $v_{i,t} > 40$, triggering a heavy penalty and episode termination.
    \item \textbf{PU Overhead:} Emptying container $i$ with volume $v$ imposes a busy time $g_i(v)$. The busy time consists of a material-dependent part that is proportional to the volume, and a constant offset for conveyor belt transport of the material from the container to the PU.
    During this period, $p_t$ counts down to zero, after which the PU is free again. Requests made while $p_t>0$ are effectively lost, emphasizing scheduling constraints.
    \item \textbf{Rewards:} The agent receives reward for emptying containers close to their peak volumes, since that behavior results in high quality output. Rewards are designed so that emptying at the higher peak is preferable. Container overflow is a terminal state with an episode reset. \emph{Invalid} or wasteful empties (e.g., container already empty or PU busy) incur a negative penalty \(\,r_{\text{pen}}\), while the \emph{do-nothing} action yields zero. 
\end{compactitem}

\paragraph{Collision state.}
A \emph{collision} arises when one or more containers simultaneously 
approach or exceed their higher-peak volumes, but the PU remains busy processing another container. 
This can rapidly push volumes over physical capacity if not addressed promptly. 
Although we first train the agent without a collision-specific mechanism, 
Section~\ref{sec:Collision_Model_and_Inference_Pipeline} discusses how 
we incorporate a collision model at inference time.

\paragraph{Objective.}
The agent must \emph{schedule empties} near either the higher or lower volume peak to maximize efficiency, while preventing overflows and avoiding collisions on the shared PU. The tension between delaying emptying for higher-volume payoffs versus frequent empties for safety and reduced collisions creates a challenging RL problem under delayed rewards and a scarce PU bottleneck.

% Prolonged time in this collision state can lead to further container fill-ups, pushing volumes dangerously close to the maximum capacity. 

\paragraph{Key Challenges}

\begin{compactitem}
    \item \textbf{Stochastic inflow \& sensor noise:} Each container’s fill rate depends on material type, density, and unpredictable external factors (e.g., time of day, seasonal fluctuations). Sensor noise further obscures volume estimates, making it difficult to predict precisely when a container will reach a target volume.
    \item \textbf{Delayed rewards \& class imbalance:} Some containers fill slowly, requiring hours of in-simulation time to reach the target volume. Consequently, episodes contain many “do nothing” steps, during which the state changes steadily but rewards from emptying remain sparse. As emptying actions are rare, reward-bearing moves are infrequent, creating skewed action distributions that challenge many reinforcement learning algorithms.
    \item \textbf{Dual-peak design \& Collisions:} Having two ideal emptying peaks per container creates a nuanced trade-off: waiting for the higher peak boosts efficiency but risks collisions and overflow if multiple containers converge while the PU is busy. Alternatively, settling too often for the lower peak increases emptying frequency, raising energy costs and volume deviations. Balancing these opposing factors—alongside scheduling constraints to avoid collisions and safety violations—poses a central challenge.
\end{compactitem}

\begin{algorithm}[H]
\caption{Three-Phase Reward Computation}
\label{alg:three_phase_reward}
\SetKwInOut{Input}{Input}
\SetKwInOut{Output}{Output}
\Input{Phase $ph \in \{1,2,3\}$, current volume $v_{t}$, action $a_t$, penalty $r_{\text{pen}}$, 
       peaks $(v_{\text{low}},v_{\text{high}})$, 
       Gaussian params $(h,w)$ or $(h_{1},h_{2},w_{1},w_{2})$}
\Output{Immediate reward $r_t$}

\BlankLine

\If(\tcp*[f]{No-op}){$a_t = 0$}{
    $r_t \leftarrow 0$\;
}
\Else{
    \If(\tcp*[f]{Invalid empty}){(\emph{invalid conditions})}{
       $r_t \leftarrow r_{\text{pen}}$\;
    }
    \Else{
       \uIf(\tcp*[f]{Phase 1}){$ph = 1$}{
         $r_t \;\leftarrow\; (h - r_{\text{pen}})\,
           \exp\!\bigl(-\tfrac{(v_t - v_{\text{high}})^2}{2\,w^2}\bigr)\;+\; r_{\text{pen}}$\;
       }
       \uElseIf(\tcp*[f]{Phase 2}){$ph = 2$}{
         $r_t \;\leftarrow\; r_{\text{pen}}
           \;+\; \sum_{i \in \{\text{low},\text{high}\}}
               \!\Bigl[h_{i} - r_{\text{pen}}\Bigr]\,
               \exp\!\bigl(-\tfrac{(v_t - v_{i})^2}{2\,w_{i}^2}\bigr)$\;
       }
       \Else(\tcp*[f]{Phase 3}){
         \uIf{$|v_{t} - v_{\text{low}}|\le 1 \;\vee\;
               |v_{t} - v_{\text{high}}|\le 1$}{
           $r_t \leftarrow 1.0$
         }
         \Else{
           $r_t \leftarrow 0$
         }
       }
    }
}
\end{algorithm}

\section{Methodology}\label{sec:methodology}

In this section, we detail our complete pipeline for training PPO-based agents to manage containers. We begin with a \emph{naive PPO} baseline, highlighting its struggles with sparse, multimodal rewards and the tendency toward premature empties. To address these issues, we then propose a \emph{curriculum learning} scheme (PPO-CL) that incrementally shapes the agent’s reward landscape. This two-stage progression—naive PPO followed by PPO-CL—lays the groundwork for an inference-time \emph{collision model}, which further mitigates overflow risks when multiple containers compete for the PU.

\subsection{Naive PPO Baseline and Its Shortcomings}

A straightforward approach is to train a PPO agent directly on the final (multimodal) reward that encourages emptying containers at either the higher or lower peak. However, as outlined in Section~\ref{sec:real_env_description}, this environment poses multiple challenges: rewards are delayed and infrequent, containers fill at varying rates, and collisions can arise when the PU services multiple containers simultaneously. Without additional structure or foresight, empirical results (see table \ref{tab:peak_ratio_actions}) show that a naive PPO agent tends to:
\begin{itemize}
    \item \textbf{Ignore long-term returns.} Driven by sparse rewards, the agent often performs multiple partial empties rather than waiting for the larger peak. This short-sighted strategy blocks the PU more frequently, increasing energy usage and leaving less capacity for containers that are about to overflow.
    \item \textbf{Struggle with skewed action distributions.} With many “do-nothing” steps before a container reaches its target volume, the agent fails to learn precise timing to consistently hit higher-volume empties.
    \item \textbf{Overlook future collisions.} Having no explicit mechanism to anticipate bottlenecks on the PU, the agent may wait too long and face simultaneous arrivals at near-peak volumes, risking overflow or forced early empties.
\end{itemize}

\subsection{Curriculum Learning with PPO}
\label{sec:curriculum_ppo}

These shortcomings motivate a more structured approach to handle delayed rewards, skewed actions, and collision risks. We therefore present a \emph{curriculum learning} strategy that gradually refines the agent’s timing and decision-making.
We train a PPO agent with a \emph{three-phase reward curriculum}, gradually introducing complexity over successive training segments. In each phase, the agent follows the same \emph{state} and \emph{action} definitions, but the reward function evolves to guide the agent toward better timing of empties:

\begin{itemize}
    \item \textbf{Phase 1 (Unimodal Reward):} We place a single Gaussian peak at the higher peak volume and no reward at the lower peak, helping the agent learn to avoid excessively early empties.
    \item \textbf{Phase 2 (Multimodal Reward):} Two Gaussian peaks (higher and lower), letting the agent discover a fallback if waiting for the higher peak is unsafe or if the PU is unavailable.
    \item \textbf{Phase 3 (Step Reward):} A strict scheme awarding positive reward only when the emptied volume is within a narrow window (\(\pm1\)) around \emph{either} peak, refining precision once the agent learned to handle both targets.
\end{itemize}

\noindent
As in {\cite{Pendyala2024-Solving,Wang2020-qc}, we further stabilize training by \emph{freezing} the policy network in parts of Phase~2, updating only the value estimator to account for changes in reward structure. During Phase~3, we \emph{unfreeze} the policy network but apply a stricter KL-divergence constraint, ensuring the agent does not deviate too aggressively from the policy learned in earlier phases. Algorithm~\ref{alg:three_phase_reward} presents the logic for all three phases. 
By stepping through these phases with carefully tuned budgets, the agent (\emph{PPO-CL}) acquires more robust scheduling behaviors than a naive single-stage approach. In particular, it learns to occasionally pick the lower peak to avert overflow. However, PPO-CL alone remains largely myopic about \emph{collisions} across containers, motivating our inference-time mitigation strategy in the next section.

\section{Collision Model and Inference Pipeline} \label{sec:Collision_Model_and_Inference_Pipeline}

Although PPO-CL improves emptying behavior, it still shows no signs of successfully learning about collision risks. We mitigate the problem by designing a mechanism to handle situations where multiple containers simultaneously approach their peak volumes. We address this by introducing a \emph{collision model (CM)} trained offline on pairwise container data, then integrating it with the PPO-CL agent at inference time to form \textbf{PPO-CL-CM}. This approach balances high-volume empties with timely overrides to avert collisions, all at minimal run-time overhead.

\paragraph{\textbf{Monte Carlo Rollouts for Pairwise Collisions:}} For efficient inference-time planning, we generate a large offline dataset of pairwise collision scenarios. By simulating isolated or small groups of containers, we capture diverse collision states with minimal run-time cost. We simulate each container pair \((i,j)\) under stochastic filling \eqref{eq:random_walk}. Two million repetitions across a container configuration yield a comprehensive offline data set of near-capacity, overflow, and collision events, minimizing deployment computation. For each pair, we perform:

\begin{enumerate}
    \item {Random initialization:} Sample means and standard deviations \(\mu_i, \mu_j, \sigma_i, \sigma_j\) for filling rates.
    \item {Stochastic evolution:}  Evolve volumes \(v_i(\tau), v_j(\tau)\) over timesteps \(\tau\).
    \item {Collision bookkeeping:}  Record collisions when both containers near peaks while the PU is busy.
\end{enumerate}

\paragraph{\textbf{Feature Extraction and Training}}

At each simulation timestep, we extract collision-predictive features:
\begin{itemize}
    \item {Volumes:} Container states \(\bigl(v_i, v_j\bigr)\)
    \item {Proximity to peaks:}  \(\Delta v_i = p_i - v_i, \Delta v_j = p_j - v_j\)
    \item {Filling parameters:} \((\mu_i, \sigma_i, \mu_j, \sigma_j)\)
    \item {Time-lag context:}  Optional short-volume histories \(\{v_i(\tau-1),v_j(\tau-1)\}\)
\end{itemize}
Each timestep receives a \emph{collision label} \(\{0,1\}\), creating a supervised dataset \(\{\text{features}, \text{collision label}\}\) for training. From this dataset, we train an XGBoost classifier to estimate collision probabilities: \(P\bigl(C_{i,j} \,\mid\, s_t, a_t\bigr) = f_{\text{col}}\!\bigl(\,\text{features}_{i,j}\bigr)\). XGBoost’s gradient-boosted trees efficiently model complex relationships between volumes, fill rates, and near-capacity states. Tuned XGBoost offers fast, accurate pairwise collision risk predictions. 

\paragraph{\textbf{Inference-Time Integration with PPO-CL:}}
Once trained, the collision model $f_{\text{col}}$ is invoked at each decision step to assess pairwise collision probabilities. The \emph{baseline} PPO-CL agent first proposes an action $a_t$. If it decides to empty a specific container ($a_t \neq 0$), we accept that choice directly. However, if PPO-CL proposes \emph{no operation} ($a_t=0$), the system queries $f_{\text{col}}$ for all container pairs to estimate near-future collision risks. If any container at high volume is flagged with a collision probability above a threshold $\theta$, we \emph{override} the no-op by forcing an empty action on the most at-risk container. Algorithm~\ref{alg:integrated_decision_making_expanded} details the procedure.

\noindent\textbf{Action Override Rationale.}
By limiting overrides to the no-op action, we minimally disturb PPO-CL’s learned preference for waiting until containers reach higher volumes. Only when the model detects a high probability of overflow or severe collisions do we \emph{force} an empty on a likely-to-clash container. This blend of \emph{offline collision modeling} and \emph{selective inference-time planning} yields a more collision-aware agent.
% without requiring complex multi-agent rollouts at run-time.

% \paragraph{\textbf{Key Contributions:}}
% \begin{compactitem}
%     \item \textbf{Offline Monte Carlo Data:} Thorough pairwise simulations capture diverse near-capacity states at low cost.
%     \item \textbf{Lightweight Collision Inference:} The model outputs collision probabilities on-demand, with no additional overhead on PPO-CL’s training loop.
%     \item \textbf{Targeted Overrides:} We only override when collisions are imminent, preserving PPO-CL’s efficiency gains elsewhere.
% \end{compactitem}

\begin{algorithm}[H]
\caption{Integrated Inference-Time Decision with Pairwise Collision Prediction}
\label{alg:integrated_decision_making_expanded}
\SetKwInOut{Input}{Input}
\SetKwInOut{Output}{Output}
\Input{State $s_t$, PPO-CL policy $\pi_{\text{CL}}$, collision model $f_{\text{col}}$, threshold $\theta$, peak volumes $\{p_i\}$}
\Output{Final action $a_{\text{final}} \in \{0,1,\ldots,n\}$}

\BlankLine
\textbf{1. PPO-CL Action:} 
$a_t \;\sim\; \pi_{\text{CL}}(a_t \mid s_t)$\;

\If{$a_t \neq 0$}{
    \Return $a_{\text{final}} \leftarrow a_t$\tcp*{Accept non-zero action}
}

\BlankLine
\textbf{2. Collision Assessment:}\\
\ForEach{pair $(i,j)$}{
  $P(C_{i,j}) \;\leftarrow\; f_{\text{col}}\!\bigl(\text{features}_{i,j}\bigr)$\;
}
Assemble matrix $P_t[i,j] \;=\; P(C_{i,j})$\;

\BlankLine
\textbf{3. Check Potential Collision Overrides:}\\
$\mathcal{C} \;\leftarrow\; \bigl\{\,i \,|\, v_i(t) \ge p_i - \delta \bigr\}$\;
\If{$\mathcal{C} \,\neq\, \emptyset$}{
   \ForEach{$i \,\in\, \mathcal{C}$}{
     $\text{CollisionRisk}(i) \;\leftarrow\;\textsc{RiskScore}(i, P_t)$
   }
   $i^\star \;\leftarrow\; \arg\max_{i \in \mathcal{C}}\! \bigl[\text{CollisionRisk}(i)\bigr]$\;
   \uIf{$\text{CollisionRisk}(i^\star) \,\ge\, \theta$}{
     $a_{\text{final}} \;\leftarrow\; i^\star$ \tcp*{Override with container $i^\star$}
   }\Else{
     $a_{\text{final}} \;\leftarrow\; 0$ \tcp*{Retain no-op}
   }
   \Return $a_{\text{final}}$
}
\BlankLine
\Return $a_{\text{final}} \leftarrow 0$\tcp*{No containers at risk, do nothing}
\end{algorithm}

% Overall, PPO-CL-CM leverages domain-aware collision prediction to mitigate overflow risks and scheduling bottlenecks, improving throughput and safety compared to naive or purely RL-driven policies.

\begin{figure*}[htbp]
    \centering
    \begin{subfigure}[b]{0.48\textwidth}
        \centering
        \includegraphics[width=\linewidth]{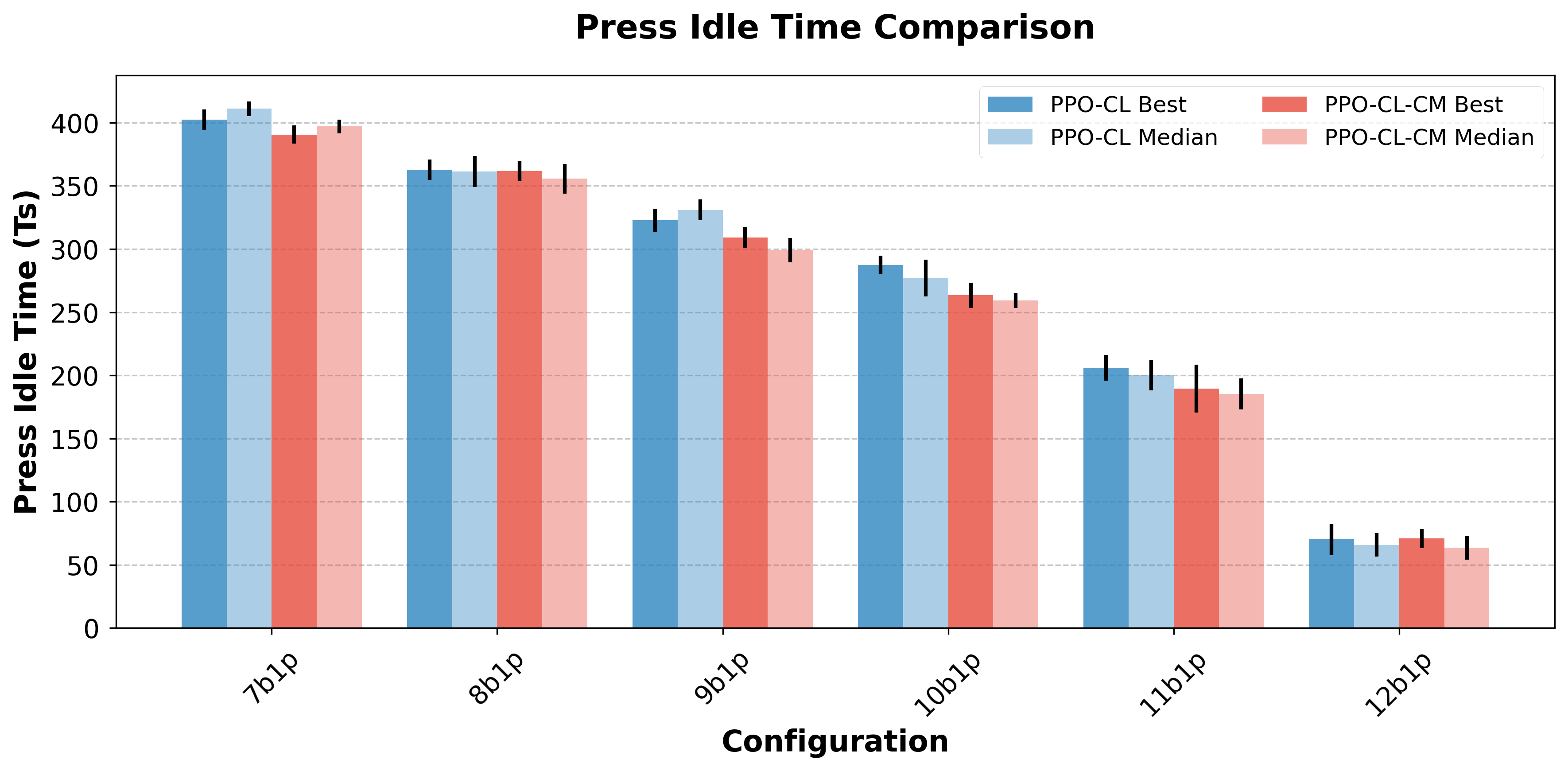}
        \caption{Press idle time comparison.}
        \label{fig:press_idle}
    \end{subfigure}
    \hfill
    \begin{subfigure}[b]{0.48\textwidth}
        \centering
        \includegraphics[width=\linewidth]{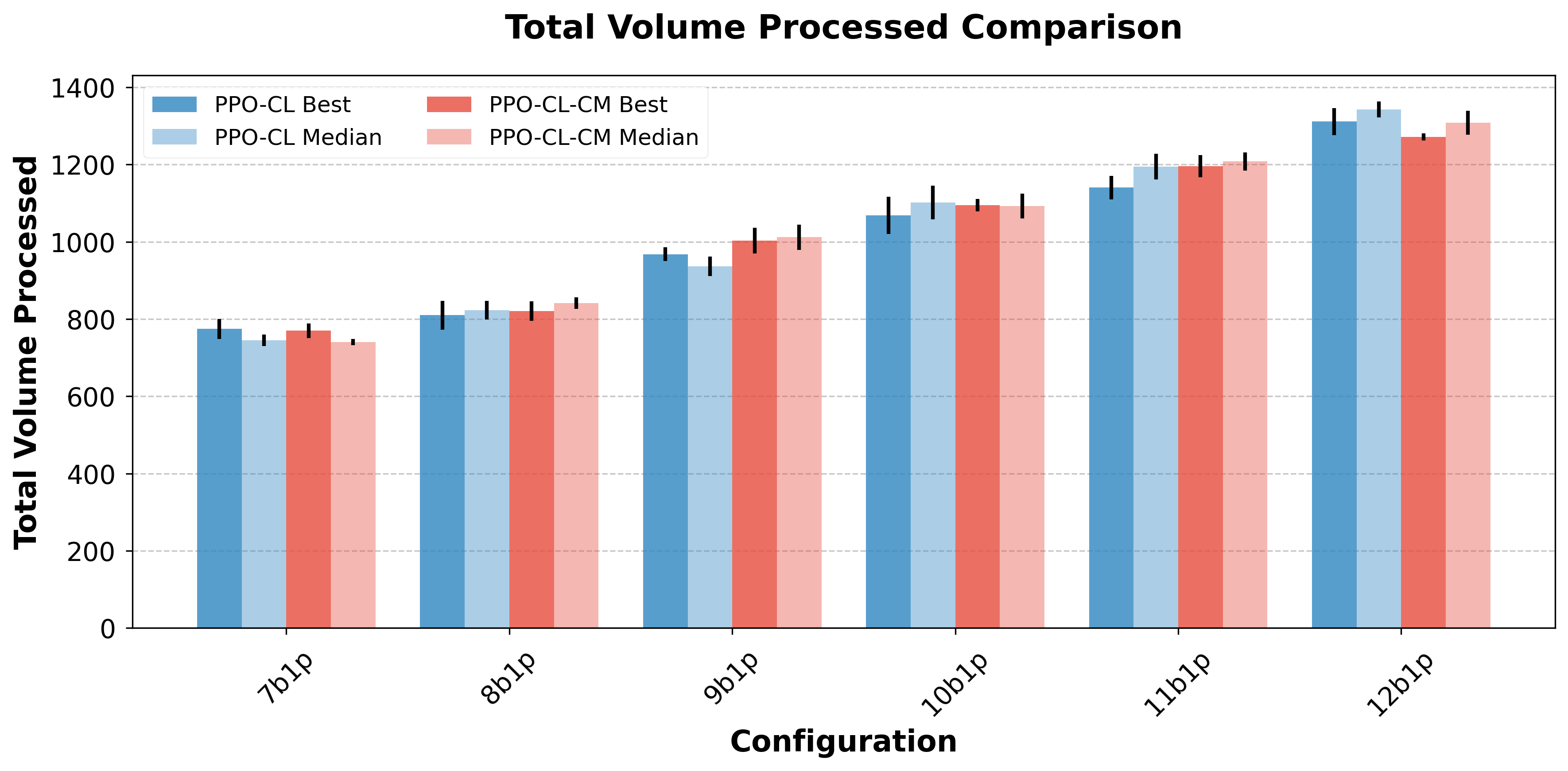}
        \caption{Total volume processed comparison.}
        \label{fig:total_volume}
    \end{subfigure}
    \caption{Performance metrics comparison between PPO-CL and PPO-CL-CM methods across different container configurations (7b1p to 12b1p). The bars show mean values and error bars indicate standard deviation. Left: Press idle time shows the duration the press remains inactive. Right: Total volume processed indicates the amount of material handled during one inference episode of 600 timesteps.}
    \label{fig:performance_comparison}
\end{figure*}

\begin{figure*}[htbp]
    \centering
    \includegraphics[width=\textwidth]{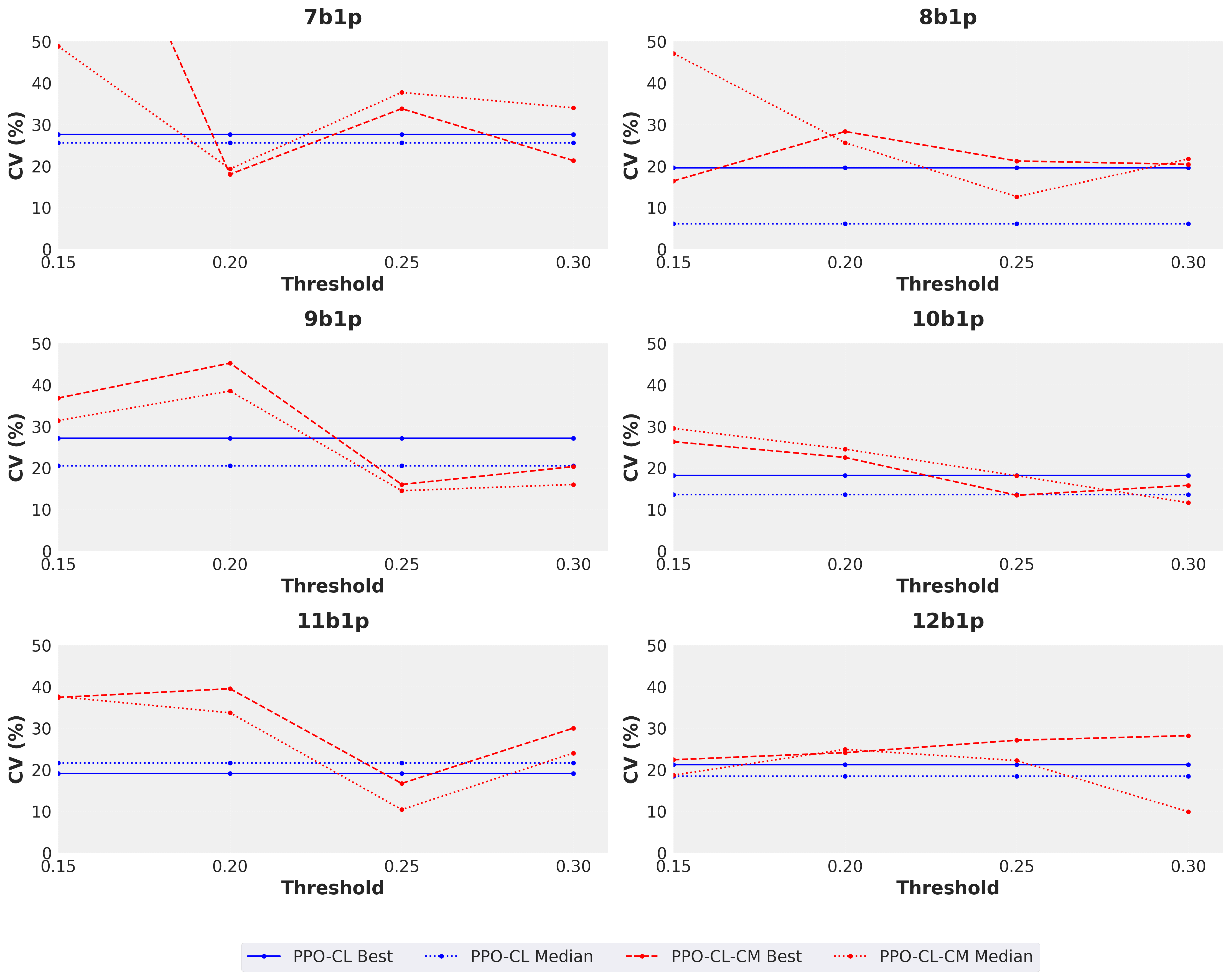}
    \caption{Comparison of Coefficient of Variation (CV\%) across different collision probability thresholds for all container configurations. Each subplot shows the performance of PPO-CL and PPO-CL-CM methods for a specific configuration. Lower CV\% indicates more consistent performance.}
    \label{fig:cv_comparison}
\end{figure*}

\section{Experimental Evaluation}\label{sec:results}

In this section, we present a quantitative evaluation of the three agents naive PPO, PPO-CL, and PPO-CL-CM across multiple container-to-PU configurations. Our analysis addresses the following research questions (\textbf{RQs}):

\begin{enumerate}
    \item \textbf{RQ1:} Is the inference-time collision model (\emph{PPO-CL-CM}) effective compared to  PPO-CL and naive PPO in terms of achieved reward?
    \item \textbf{RQ2:} How effectively does the collision model reduce safety-limit violations (i.e., empties above the critical volume limit)?
    \item \textbf{RQ3:} How do these findings inform real-world design choices, such as the ideal ratio of containers to processing units?
\end{enumerate}

In the spirit of open and reproducible research, we make our source code available via an \textbf{\emph{anonymous repository}}.%
\footnote{\url{https://gitlab.com/anonymousppocl_cm1/anonymous_collisions_paper}}
The repository contains a script for reproducing all results presented in this section.

\subsection{Experimental Setup}

We evaluate our methods on environments with \emph{7 to 12} containers and a single PU (labeled ``7b1p'' through ``12b1p''). Following \cite{Pendyala2024-ContainerGym} we use an episode length of 600 timesteps, with a 60-second granularity. 
For each agent, we conduct \emph{15} independent training runs using distinct random seeds. 
During inference, we run each seed-based policy in \emph{5} rollouts and collect statistics. 
We present results for both the \emph{best} and \emph{median} performing seeds of each method, ensuring a fair and comprehensive comparison. Specifically, the \textit{best seed} is selected based on the smallest number of \emph{collision timesteps}, while the \textit{median seed} is chosen from the remaining seeds in terms of the same metric.

\paragraph{Metrics:}
The reward signal aggregates many different subgoals. To obtain a fine-grained picture of algorithm performance, we monitor the following performance metrics:
\begin{itemize}
    \item \textit{Press Idle Time} (\figurename~\ref{fig:press_idle}): total timesteps during which the PU is not processing any container.
    \item \textit{Total Volume Processed} (\figurename~\ref{fig:total_volume}): material throughput in one 600-step inference episode.
    \item \textit{Collisions in time-steps}: the total number of timesteps in which a \emph{collision state} occurs; that is, at least two containers simultaneously approach or exceed their ideal volumes (especially the higher peak) while the PU is busy and thus unable to service them.
    \item \textit{Coefficient of Variation (CV\%) (\figurename~\ref{fig:cv_comparison})}: a normalized measure of variability in performance under different collision thresholds. 
    \item \textit{Total Volume Deviation}: the average (over all containers and timesteps) of the absolute difference between a container’s actual volume and its nearest ideal peak, gauging how well empties align with target volumes.
    \item \textit{Actions per Container, Reward per Action, and Peak-Usage Ratios}: drawn from Tables~\ref{tab:main_metrics} and~\ref{tab:peak_ratio_actions}.
    \item \textit{Safety-Limit Violation Percentage} (\figurename~\ref{fig:safety_limit_violation_percentage}): the fraction of emptying actions during inference where a container volume exceeds a fixed \emph{\textbf{critical volume limit}}, set 5~units above that container’s higher ideal emptying peak.
\end{itemize}

\noindent All agents are tested under the same environmental conditions to isolate the impact of curriculum training and collision modeling. 
Moreover, the results shown in Figures{~\ref{fig:performance_comparison}} and{~\ref{fig:safety_limit_violation_percentage}} and in Tables{~\ref{tab:main_metrics}} and{~\ref{tab:peak_ratio_actions}} are reported using the threshold values that yield the lowest CV\% in Figure{~\ref{fig:cv_comparison}}, reflecting the most collision-stable configurations in our analysis.

\subsection{Impact of Collision Model (RQ1)}

From Table~\ref{tab:peak_ratio_actions}, we note that naive PPO often fails to empty containers at the \emph{higher} ideal peak altogether (\emph{e.g.}, ratio close to zero), indicating it resorts to early or sub-optimal empties. This behavior leads to more frequent episodes ending prematurely due to overflow (especially in the larger ``12b1p'' setup) or consistently high volume deviations. While PPO-CL capitalizes on the multi-phase reward shaping to handle delayed feedback and class imbalance, collisions can still occur if multiple containers simultaneously converge on their higher peaks. This is where incorporating \emph{inference-time collision checks} to yield PPO-CL-CM has an edge. 

As shown in \figurename~\ref{fig:cv_comparison}, PPO-CL-CM achieves lower variability (CV\%) across different collision probability thresholds, meaning it more consistently avoids hazardous states. From \figurename~\ref{fig:press_idle}, we see PPO-CL-CM generally reduces press idle time compared to PPO-CL, indicating fewer deadlocks where containers are left unemptied until near-overflow conditions. Meanwhile, \figurename~\ref{fig:total_volume} shows that \emph{total volume processed} remains at least on par with (and often surpasses) PPO-CL, demonstrating that collision avoidance does not compromise overall throughput in an inference episode. 

Table~\ref{tab:main_metrics} reveals that \emph{collision timesteps} drop systematically for \emph{PPO-CL-CM}, and its \emph{volume deviation} is also slightly lower on average. The \emph{higher/lower peak ratio} in Table~\ref{tab:peak_ratio_actions} confirms that PPO-CL-CM empties containers earlier (lower peak) only when the collision model flags imminent risk, thereby balancing high-throughput empties with safety. Hence, \textbf{RQ1} is answered: adding an inference-time collision model mitigates bottlenecks and collisions beyond what curriculum-based RL can achieve alone.

\subsection{Safety-Limit Violations (RQ2)}
\label{subsec:safety_limit_violations}

Figure~\ref{fig:safety_limit_violation_percentage} summarizes the frequency of empties that exceed the safety critical volume limit, thus posing a higher risk of overflow and breach of physical limit. We observe that \textbf{PPO-CL-CM} consistently maintains a smaller fraction of risky empties overall than PPO-CL for all bunker-to-PU setups from \emph{7b1p} through \emph{12b1p}. This indicates that the collision model not only reduces direct collisions but also prompts timely empties before containers venture into risky volume ranges. Hence, we conclude \textbf{RQ2} by confirming that inference-time collision checks can effectively mitigate dangerous critical empties, supporting safer operation without sacrificing throughput.

\subsection{Real-World Implications and Design Guidance (RQ3)}

\figurename~\ref{fig:press_idle} illustrates that as we move from ``7b1p'' up to ``12b1p,'' \emph{press idle time} diminishes significantly, reflecting how the PU becomes fully utilized with rising container counts. On the other hand, in extremely large configurations (e.g., 12 containers to 1 PU), collisions inevitably remain because a single resource cannot realistically handle multiple near-peak arrivals at once. While PPO-CL-CM can curb severe collisions, it cannot eliminate them entirely when the resource ratio is unfavorable. Thus for \emph{Moderate Ratios:} Up to around 7--11 containers per PU, collision avoidance measures like PPO-CL-CM yield strong improvements without saturating the system. But for \emph{High Ratios:}, beyond a certain limit (e.g., 12b1p), press idle time becomes negligible, and collision events dominate. An additional PU may be necessary for higher container configurations. Addressing \textbf{RQ3}, in practice, operators can apply these findings when deciding how many containers can be feasibly connected to a single processing unit, and whether advanced collision checks are cost-effective. Importantly, applying a collision model allows to safely operate more containers with the same number of expensive PUs.

\begin{table}[htbp]
    \centering
    \footnotesize %\scriptsize 
    \setlength{\tabcolsep}{2.5pt}  % Slightly reduce column spacing
    \begin{tabular}{@{}lcccc@{}}
    \toprule
    \textbf{Config} & \textbf{Agent} 
                    & \textbf{Collisions (Ts)} 
                    & \textbf{Tot.\ vol.\ dev (\%)} 
                    & \textbf{Reward/action} \\
    \midrule
    
    \multirow{2}{*}{7b1p}
      & PPO-CL 
        & 72.4$\pm$20.0 / 81.6$\pm$20.9
        & 7.3$\pm$2.1 / 6.9$\pm$1.2
        & 0.4 / 0.4 \\
      & PPO-CL-CM 
        & \textbf{22.0$\pm$3.9} / \textbf{34.4$\pm$6.7}
        & \textbf{5.1$\pm$0.7} / \textbf{6.1$\pm$0.6}
        & \textbf{0.7} / \textbf{0.8} \\
    \midrule
    
    \multirow{2}{*}{8b1p}
      & PPO-CL
        & 77.2$\pm$15.1 / 98.6$\pm$6.1
        & 7.4$\pm$0.8 / 7.3$\pm$0.9
        & 0.3 / 0.4 \\
      & PPO-CL-CM
        & \textbf{66.2$\pm$14.0} / \textbf{80.6$\pm$10.2}
        & \textbf{6.7$\pm$1.2} / \textbf{7.0$\pm$0.7}
        & \textbf{0.6} / \textbf{0.5} \\
    \midrule
    
    \multirow{2}{*}{9b1p}
      & PPO-CL
        & 117.0$\pm$31.7 / 153.4$\pm$31.4
        & 6.5$\pm$1.4 / 7.5$\pm$1.1
        & 0.4 / 0.3 \\
      & PPO-CL-CM
        & \textbf{89.8$\pm$14.4} / \textbf{117.0$\pm$17.0}
        & \textbf{6.3$\pm$1.4} / 7.6$\pm$0.7
        & \textbf{0.6} / \textbf{0.4} \\
    \midrule
    
    \multirow{2}{*}{10b1p}
      & PPO-CL
        & 154.6$\pm$28.1 / 189.8$\pm$25.8
        & 6.8$\pm$1.2 / 8.7$\pm$0.9
        & 0.4 / 0.2 \\
      & PPO-CL-CM
        & \textbf{138.2$\pm$18.5} / \textbf{145.4$\pm$26.3}
        & \textbf{6.4$\pm$0.4} / \textbf{6.8$\pm$0.9}
        & \textbf{0.5} / \textbf{0.5} \\
    \midrule
    
    \multirow{2}{*}{11b1p}
      & PPO-CL
        & 130.8$\pm$25.0 / 156.4$\pm$33.8
        & 8.0$\pm$1.3 / 6.9$\pm$0.5
        & 0.3 / 0.3 \\
      & PPO-CL-CM
        & \textbf{83.2$\pm$13.9} / \textbf{148.4$\pm$15.4}
        & \textbf{7.6$\pm$1.2} / \textbf{6.8$\pm$0.7}
        & \textbf{0.5} / \textbf{0.5} \\
    \midrule
    
    \multirow{2}{*}{12b1p}
      & PPO-CL
        & 126.4$\pm$26.8 / 195.8$\pm$36.0
        & 11.8$\pm$2.1 / 10.2$\pm$1.3
        & 0.3 / 0.3 \\
      & PPO-CL-CM
        & \textbf{117.4$\pm$31.9} / \textbf{175.6$\pm$38.9}
        & 13.0$\pm$2.2 / 11.2$\pm$1.6
        & \textbf{0.5} / \textbf{0.4} \\
    \bottomrule
    \end{tabular}
    \caption{%
      Main performance metrics (Best/Median) for each bunker configuration, 
      all rounded to one decimal place. Each cell shows 
      \emph{Best} $\pm$ std.\ / \emph{Median} $\pm$ std.\ in one line. 
      We boldface the \textbf{better} result in PPO-CL-CM whenever 
      it outperforms PPO-CL (lower collisions/dev or higher reward).}
    \label{tab:main_metrics}
\end{table}

% % Table 2: Peak Metrics
\begin{table}[htbp]
    \centering
    \footnotesize %scriptsize
    \setlength{\tabcolsep}{2pt}  % Adjust column spacing as needed
    \begin{tabular}{@{}llcccccc@{}}
    \toprule
    \multirow{2}{*}{Config}
     & \multirow{2}{*}{Type}
     & \multicolumn{3}{c}{Higher/Lower Peak \% Ratio}
     & \multicolumn{3}{c}{Actions/Container} \\
    \cmidrule(lr){3-5}\cmidrule(lr){6-8}
    & & {\scriptsize \textbf{Naive PPO}} & {\scriptsize \textbf{PPO-CL}} & {\scriptsize \textbf{PPO-CL-CM}}
      & {\scriptsize \textbf{Naive PPO}} & {\scriptsize \textbf{PPO-CL}} & {\scriptsize \textbf{PPO-CL-CM}} \\
    \midrule

    \multirow{2}{*}{7b1p}
      & Best   
        & 0.0 & 45.7 & 1.0 
        & 32.7$\pm$8.9 & 20.0$\pm$3.4 & 25.6$\pm$2.6 \\
      & Median 
        & 0.0 & 18.6 & 0.8 
        & 29.6$\pm$12.3 & 19.6$\pm$2.8 & 24.4$\pm$3.1 \\
    \midrule

    \multirow{2}{*}{8b1p}
      & Best   
        & 0.0 & 8.4 & 2.2 
        & 31.0$\pm$10.0 & 18.9$\pm$3.8 & 20.8$\pm$3.9 \\
      & Median 
        & 0.0 & 14.0 & 2.9 
        & 31.2$\pm$8.6 & 18.8$\pm$3.1 & 20.9$\pm$4.7 \\
    \midrule

    \multirow{2}{*}{9b1p}
      & Best   
        & 0.0 & 16.5 & 4.5 
        & 33.3$\pm$9.0 & 19.4$\pm$3.4 & 21.2$\pm$2.7 \\
      & Median 
        & 0.0 & 19.6 & 3.3 
        & 34.8$\pm$9.9 & 18.3$\pm$4.1 & 21.4$\pm$3.1 \\
    \midrule

    \multirow{2}{*}{10b1p}
      & Best   
        & 0.0 & 30.5 & 3.3 
        & 33.2$\pm$10.3 & 18.9$\pm$2.5 & 20.8$\pm$4.2 \\
      & Median 
        & 0.0 & 11.7 & 3.1 
        & 31.6$\pm$8.0 & 19.1$\pm$2.8 & 20.7$\pm$5.0 \\
    \midrule

    \multirow{2}{*}{11b1p}
      & Best   
        & 0.0 & 3.1 & 1.8 
        & 34.1$\pm$6.3 & 20.6$\pm$4.6 & 22.5$\pm$4.3 \\
      & Median 
        & 0.0 & 6.7 & 2.9 
        & 32.9$\pm$7.6 & 20.3$\pm$2.7 & 21.7$\pm$3.5 \\
    \midrule

    \multirow{2}{*}{12b1p}
      & Best   
        & N/A & 1.9 & 1.6 
        & N/A & 23.1$\pm$7.0 & 23.2$\pm$8.1 \\
      & Median 
        & N/A & 2.6 & 2.4 
        & N/A & 22.1$\pm$4.8 & 22.1$\pm$6.1 \\
    \bottomrule
    \end{tabular}
    \caption{%
      Comparison of Higher/Lower Peak Percentage Ratio and Mean Actions per container 
      (single decimal precision). For each configuration, ``Best'' / ``Median'' rows 
      show mean $\pm$ standard deviation where applicable.
    }
    \label{tab:peak_ratio_actions}
\end{table}

% Safety limit violation percentage figure
\begin{figure}[htbp]
    \centering
    \includegraphics[width=0.9\textwidth]{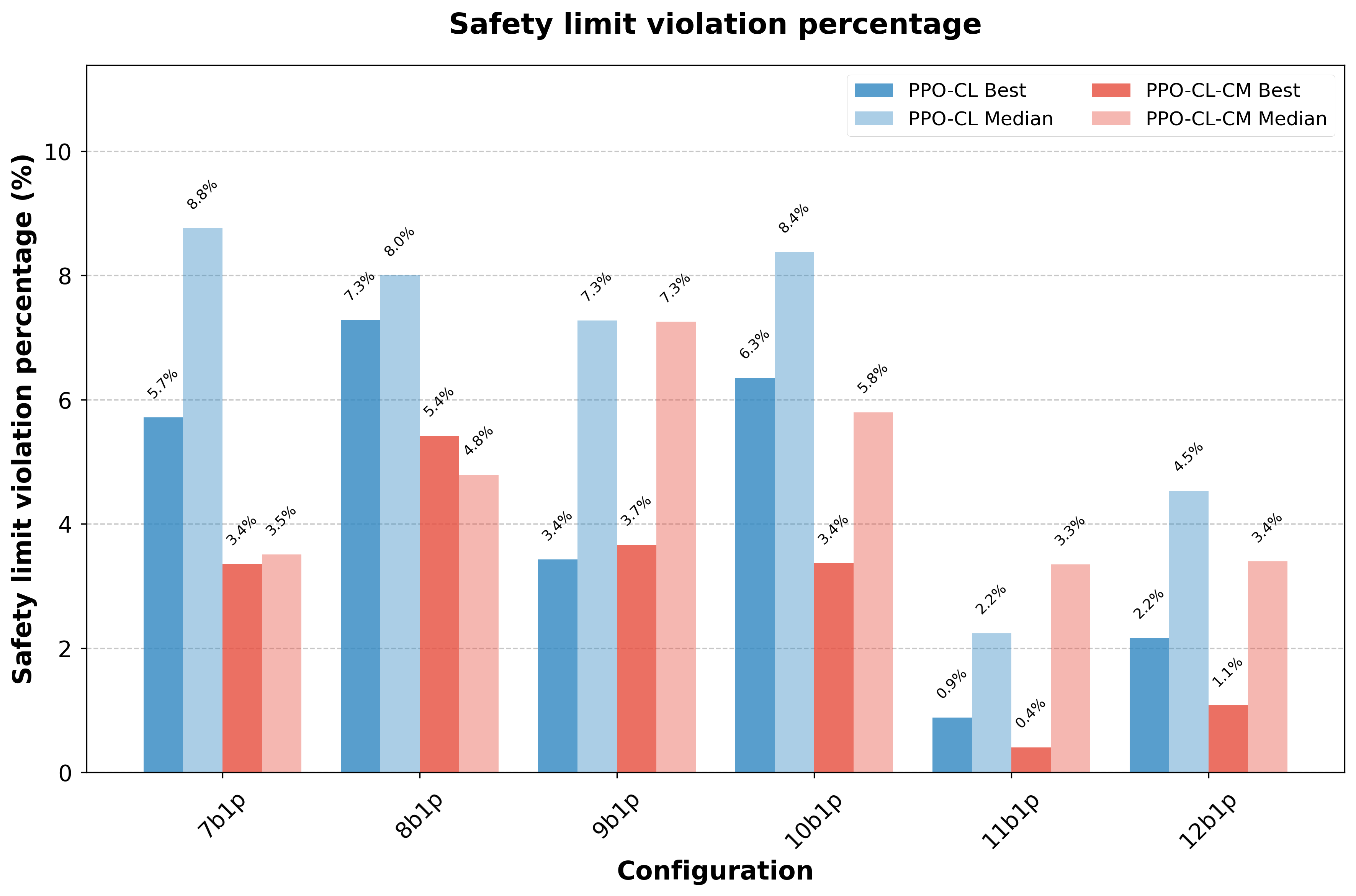}
    \caption{Comparison of safety limit violation percentages across different bunker configurations. Bars show the percentage of emptying actions that exceeded the safety limit for each bunker configuration using PPO-CL and PPO-CL-CM methods.}
    \label{fig:safety_limit_violation_percentage}
\end{figure}

\section{Conclusion}\label{sec:conclusion}

We have presented a hybrid strategy for container management in a waste-sorting facility in a constrained resource scenario, where each container has two preferred ``ideal'' emptying volumes. The first component, \emph{PPO-CL}, addresses the challenges of sparse rewards and dual-volume targets through a curriculum-learning approach that teaches the agent to empty containers near either a lower or higher peak. The second component, a \emph{collision model} integrated at inference time (\emph{PPO-CL-CM}), provides targeted overrides when containers risk colliding at peak volumes.  This combination effectively balances the key design criteria: prioritizing the higher peak for optimal throughput while resorting to the lower peak only when collisions or safety violations become imminent.

Empirical results across various container-to-PU ratios (7:1 to 12:1)  demonstrate that \emph{PPO-CL-CM} reduces both collision episodes and empties above a critical safety threshold, \emph{without sacrificing total processed volume}. By explicitly modeling future collision risk, it prevents premature empties that might otherwise occur if the agent tried to avoid collisions by abandoning the higher peak too soon. From an operational standpoint, these findings suggest that facility managers can confidently scale up container counts to a point, relying on our framework to avoid overflows and to ensure safe, high-volume empties. Beyond that point, collisions become unavoidable, but our method still lessens their severity.

A notable advantage of this framework is that the collision model is trained entirely offline via Monte Carlo simulations, adding only minimal overhead during inference. This approach stands in contrast to computationally intensive online planners like Monte Carlo Tree Search, making it better suited for large, stochastic industrial environments with real-time decision needs. Looking ahead, we envision several avenues to extend this work: integrating multiple PUs (or more complex resource constraints) within the same collision-avoidance framework, dynamically tuning collision thresholds based on time-of-day inflows or real-time capacity data, and more tightly fusing offline collision insights with online RL updates to further refine scheduling decisions. Our findings highlight that a domain-aware collision model, combined with carefully shaped RL curricula, can yield safer, more efficient container management, establishing a robust framework for broader industrial adoption and more complex resource-allocation tasks.

\subsubsection{Acknowledgements: }

This work was funded by the German federal ministry of economic affairs and climate action through the ``ecoKI'' grant.

% \bibliography{bibliography}

\end{document}